\documentclass{article}

\usepackage{microtype}
\usepackage{graphicx}
\usepackage{subfigure}
\usepackage{booktabs} 

\usepackage{hyperref}

\usepackage[accepted]{sysml2019}

\DeclareFixedFont{\ttb}{T1}{txtt}{bx}{n}{7} 
\DeclareFixedFont{\ttm}{T1}{txtt}{m}{n}{7}  

\usepackage{color}
\definecolor{lightblue}{RGB}{17,118,197}
\definecolor{darkblue}{RGB}{20,73,111}
\definecolor{purple}{RGB}{83,70,167}
\definecolor{green}{RGB}{25,166,172}

\usepackage{amsfonts}
\usepackage{amsmath}
\usepackage{listings}

\newcommand\pythonstyle{\lstset{
language=Python,
basicstyle=\ttm,
otherkeywords={self},             
keywordstyle=\ttb\color{darkblue},
emph={MyClass,__init__},          
commentstyle=\ttm\color{lightblue},
emphstyle=\ttb\color{darkblue},    
stringstyle=\color{green},
showstringspaces=false,            %
literate={\ \ }{{\ }}1
}}

\lstnewenvironment{python}[1][]
{
\pythonstyle
\lstset{#1}
}
{}

\sysmltitlerunning{TF-Replicator: Distributed machine learning for researchers}

\begin{document}

\twocolumn[
\sysmltitle{TF-Replicator: Distributed machine learning for researchers}



\sysmlsetsymbol{equal}{*}

\begin{sysmlauthorlist}
\sysmlauthor{Peter Buchlovsky\textsuperscript{*}}{deepmind}
\sysmlauthor{David Budden\textsuperscript{*}}{deepmind}
\sysmlauthor{Dominik Grewe\textsuperscript{*}}{deepmind}
\sysmlauthor{Chris Jones\textsuperscript{*}}{deepmind}
\sysmlauthor{John Aslanides}{deepmind}
\sysmlauthor{Frederic Besse}{deepmind}
\sysmlauthor{Andy Brock}{deepmind}
\sysmlauthor{Aidan Clark}{deepmind}
\sysmlauthor{Sergio G\'omez Colmenarejo}{deepmind}
\sysmlauthor{Aedan Pope}{deepmind}
\sysmlauthor{Fabio Viola}{deepmind}
\sysmlauthor{Dan Belov}{deepmind}
\end{sysmlauthorlist}

\sysmlaffiliation{deepmind}{DeepMind, London, UK}

\sysmlcorrespondingauthor{David Budden}{budden@google.com}

\sysmlkeywords{Machine Learning, SysML}

\vskip 0.3in

\begin{abstract}
We describe TF-Replicator, a framework for distributed machine learning designed for DeepMind researchers and implemented as an abstraction over TensorFlow. TF-Replicator simplifies writing data-parallel and model-parallel research code. The same models can be effortlessly deployed to different cluster architectures (i.e. one or many machines containing CPUs, GPUs or TPU accelerators) using synchronous or asynchronous training regimes. To demonstrate the generality and scalability of TF-Replicator, we implement and benchmark three very different models: (1) A ResNet-50 for ImageNet classification, (2) a SN-GAN for class-conditional ImageNet image generation, and (3) a D4PG reinforcement learning agent for continuous control. Our results show strong scalability performance without demanding any distributed systems expertise of the user. The TF-Replicator programming model will be open-sourced as part of TensorFlow 2.0 (see https://github.com/tensorflow/community/pull/25).
\end{abstract}
]



\printAffiliationsAndNotice{}  

\section{Introduction}
\label{sec:intro}
\cite{openai_blog} demonstrated that the amount of compute used for the largest machine learning training runs has increased exponentially with a 3.5 month-doubling period since 2012 (compared to 18 months historically for Moore's Law). Although there have been noteworthy improvements in the vertical scalability of machine learning systems (e.g. by leveraging GPUs, and recently the introduction of custom accelerators such as Google's TPU~\cite{jouppi2017datacenter}), researchers have been forced to scale horizontally across massively distributed compute clusters to conduct experiments of this magnitude.

Researchers faced with building a distributed system are currently required to make a steep trade-off between simplicity and generality. High-level abstractions have been developed to simplify common use-cases, e.g. the parameter server model for synchronous SGD as supported by TensorFlow's combination of \texttt{replica\_device\_setter} and \texttt{SyncReplicasOptimizer}. Horovod~\cite{sergeev2018horovod} supports similar applications by spawning multiple data-parallel TensorFlow graphs, using MPI or NCCL \texttt{all\_reduce} implementations for efficient communication of gradients~\cite{nccl}. The TensorFlow Estimator was developed to address certain models (e.g. generative adversarial networks (GANs)~\cite{goodfellow2014generative}) in a device-agnostic manner, but deny researchers the freedom of naturally defining their own run loops.

The number of models not addressed by these abstractions is growing rapidly. In the original TensorFlow paper, the authors cited multi-loss methods and reinforcement learning (RL) agents as poorly suited to the parameter server-based model of its DistBelief predecessor~\cite{dean2012large}. Since TensorFlow's release, multi-loss methods have proliferated, including proximal gradient TD-learning~\cite{liu2016proximal}, multi-level optimization~\cite{pfau2016connecting}, synthetic gradients~\cite{jaderberg2016decoupled} and RL agents that interact with multiple losses in increasingly complex ways (e.g. hierarchical RL~\cite{vezhnevets2017feudal}). Many popular machine learning frameworks provide low-level primitives that can in principle be used to implement such systems, but their usage requires a deep understanding of distributed system architectures. 

In this paper we present TF-Replicator, designed and developed in close collaboration with DeepMind researchers across several machine learning disciplines. In Section~\ref{sec:distributedtf}, we provide an overview of the low-level mechanisms and common patterns for distributed TensorFlow, which are recurring concepts throughout the remainder of the paper. In Section~\ref{sec:replicator} we provide an overview of TF-Replicator. Starting with the API used to define a replica, we demonstrate painless generalization across different system architectures and devices, and show optional arguments to configure more complex use-cases (e.g. combining data and model parallelism). Section~\ref{sec:implementation} provides an overview of TF-Replicator's implementation details. The remainder of the paper demonstrates how TF-Replicator can be applied to achieve compelling, scalable performance for various machine learning models in both GPU and TPU clusters.

\section{TensorFlow}
\label{sec:distributedtf}
We start with a brief description of the underlying framework. TensorFlow is a computation-graph based machine learning framework, wherein a user builds a graph consisting of tensors, representing multidimensional arrays of data; and operations (ops), which consume and produce tensors. A user can request to see the value of a tensor, entailing the execution of any ancestor operations.

\subsection{Distributed TensorFlow}
\label{ss:distributed}
Each machine in a distributed TensorFlow cluster launches a TensorFlow server, communicating over gRPC. The server exports two RPC services: the \textbf{master}, which coordinates and provides access to a set of distributed machines; and the \textbf{worker}, responsible for executing subgraphs using its local devices. On one or more machines, a \textbf{client} binary (usually a Python script) defines a TensorFlow graph. Subgraphs are dispatched by the master to the workers for execution. One master can dispatch work to multiple workers, and one worker can perform work for multiple masters (e.g. the parameter server model~\cite{li2014scaling}, whereby parameter server workers store and update variable weights accessible from many parallel replicas).

During graph construction, device strings are used to identify devices (e.g. a specific GPU) in a cluster. When device ``A" requires the result of an operation with device string ``B", its master sends a gRPC to device ``B" asking its worker to run the operation. During preprocessing, TensorFlow identifies all tensors produced on one device but consumed by an operation on another, and inserts \texttt{send} and \texttt{recv} nodes to transfer data between devices. The details depend on the devices in question, for example \texttt{cudaMemcpyAsync} for local CPU $\leftrightarrow$ GPU and gRPC over TCP in the distributed setting~\cite{abadi2016tensorflow}. Relative device strings are mapped to physical network addresses in a \texttt{ClusterSpec} dictionary provided to each server at initialization.

\begin{python}
# Distributed TensorFlow with device strings.
with tf.device('/job:ps/task:0'):
    w = tf.Variable(...)
    b = tf.Variable(...)
with tf.device('/job:worker/task:0'):
    inputs = ...
    y = tf.nn.relu(tf.matmul(inputs, w) + b)
\end{python}

So that two clients can share state, TensorFlow defines \textbf{resources} (e.g. variables), objects with a unique name in addition to their device string. If two clients define a resource with identical device and name, only a single resource is allocated on the relevant device, and referring operations from either client will share the underlying data.

\subsection{Graph Replicas}
\label{ss:replicas}
\begin{figure}[t]
\label{fig:replication}
\centering
\includegraphics[width=8.5cm]{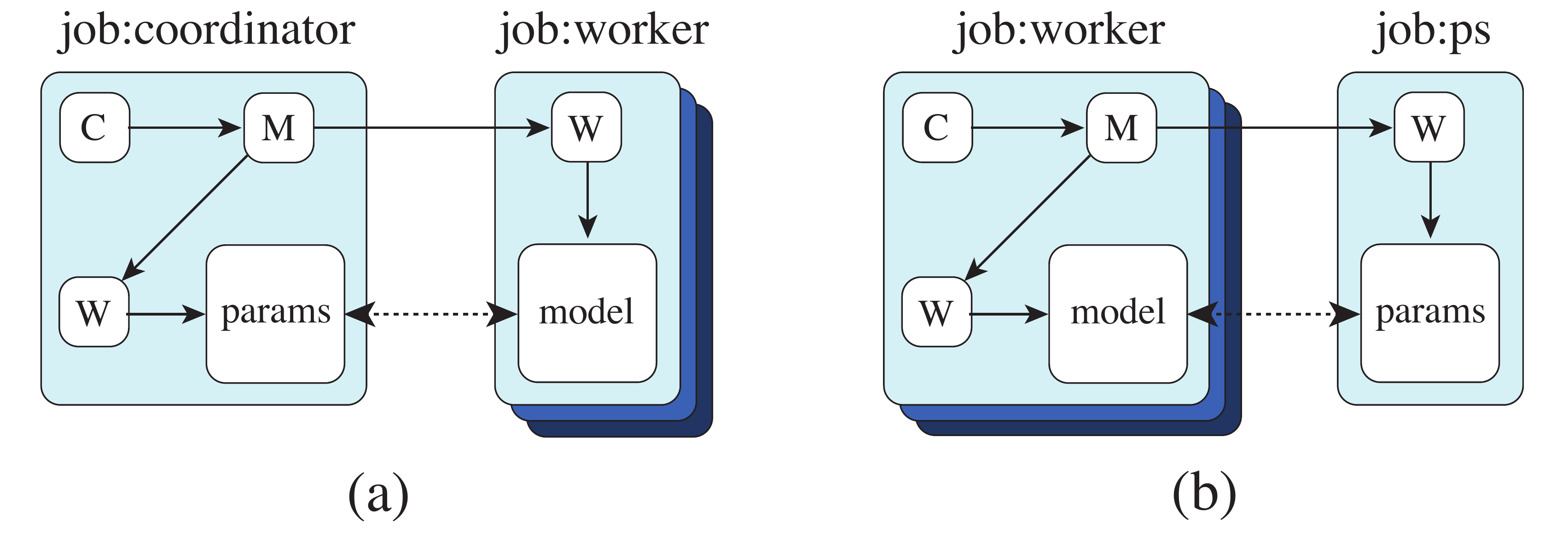}
\vspace{-2em}
\caption{Replication patterns for data parallelism in distributed TensorFlow: (a) in-graph replication, and (b) between-graph replication. C = client, M = master service, W = worker service.}
\end{figure}
TensorFlow provides a versatile platform for distributed computing with a high degree of freedom, but little abstraction. To capture common use cases, we shall use the key concept of a \textbf{replica}: a computation designed to be run in parallel across many devices (e.g. one step of SGD), with different input to each device, and some shared state accessible to all replicas (e.g. model parameters). Many machine learning techniques (e.g. Hogwild!~\cite{recht2011hogwild}, Downpour SGD~\cite{dean2012large} and A3C~\cite{mnih2016asynchronous}) can be seen as computations involving many replicas with differing synchronicity of the shared parameters. There are many ways to construct a system with replicas, but two common patterns are in-graph replication and between-graph replication (illustrated in Figure 1).

For \textbf{in-graph replication} (Figure 1(a)), a single client binary constructs one replica on each device (all contained in a massive multi-device graph), each sharing a set of parameters (resources) placed on the client's own device. That client’s master service then coordinates synchronous execution of all replicas: feeding data and fetching outputs. 

For \textbf{between-graph replication} (Figure 1(b)), multiple machines run a client binary, each creating a graph with a single replica on a local device, alongside resources on a shared device (i.e. the parameter server model). Because this relies on TensorFlow’s name resolution to share state, care must be taken to ensure each client creates resources identically. 

\subsection{Limitations}
\label{ss:limitations}
Though TensorFlow supports near-arbitrary replication, implementation is often difficult. Handling state is non-trivial, with incorrect behavior often leading to silent error. Coordinating heterogeneous behavior among replicas (not yet described), necessary for many methods in reinforcement learning, adds even more complexity. 

Google recently unveiled its TPU v2 devices, each with 180 teraflops and 64 GB of high bandwidth memory across four chips of two cores. These devices can be connected in a 2-dimensional toroidal mesh network to form a 11.5 petaflop TPU ``Pod"~\cite{cloud_tpu}. Leveraging the power of these devices demands a specific replication pattern, and requires the user to manually handle asynchronous data transfer using \texttt{InfeedQueue} and \texttt{OutfeedQueue} ops. Building an efficient system within these constraints adds further complexity to distributed TensorFlow, and was a key influence in our decision to build a new abstraction for replicated machine learning computation.

\section{TF-Replicator}
\label{sec:replicator}
TF-Replicator trivializes the process of building distributed machine learning systems by allowing researchers to naturally define their model and run loop as per the single-machine setting. This is achieved by providing a simple API for defining a (potentially model-parallel) replica, and a set of in-built Replicators that support deployment across different system architectures (in-graph versus between-graph), training regimes (synchronous versus asynchronous) and hardware (GPU and TPU). Unlike TensorFlow Estimator, it makes few assumptions about model structure and is therefore readily extensible to novel use cases.

In TF-Replicator, a replica is defined by two functions: \texttt{input\_fn} for describing how a replica receives its input, and \texttt{step\_fn} for describing the computation performed by the replica (detailed in Section~\ref{ss:defining}). To deploy a replica, the user builds one of the in-built Replicator implementations detailed in Table~\ref{table:replicators}. Any resource constructed within the Replicator context is itself replicated, e.g. the model and optimizer dataflow graphs. The Replicator \texttt{wrap\_optimizer} function coordinates variable updates by \texttt{all\_reduce} averaging of the computed gradients. Note that TF-Replicator exposes several of these MPI-style primitives for scenarios requiring inter-replica communication, as detailed in Section~\ref{ss:advanced}. Bringing the features together, a user might deploy an ImageNet classifier on a single TPU v2 (8 cores across 4 chips) as follows:

\begin{python}
# Deploying a model with TpuReplicator.
repl = tf_replicator.TpuReplicator(
    num_workers=1, num_tpu_cores_per_worker=8
)
with repl.context():
    model = resnet_model()
    base_optimizer = tf.train.AdamOptimizer()
    optimizer = repl.wrap_optimizer(base_optimizer)
# ... code to define replica input_fn and step_fn.
per_replica_loss = repl.run(step_fn, input_fn)
train_op = tf.reduce_mean(per_replica_loss)

with tf.train.MonitoredSession() as session:
    repl.init(session)
    for i in xrange(num_train_steps):
        session.run(train_op)
    repl.shutdown(session)
\end{python}


\subsection{Deploying Replicas}
\label{ss:deployment}
A summary of currently supported replica implementations is provided in Table~\ref{table:replicators}. Maximum workers refers to the number of separate machines available to the distributed system, e.g. a server with many GPUs or a TPU device constituting 8 cores split across 4 chips. Devices per worker refers to the number of physical devices available on each of these workers, e.g. each GPU on a server. Note that for the \texttt{MultiGpuReplicator}, \texttt{MultiWorkerReplicator} and \texttt{TpuReplicator}, the in-graph replication pattern is adopted for synchronous training. 

Machine learning workloads using asynchronous SGD are still common-place, as popularized by Hogwild!~\cite{recht2011hogwild}. In addition to the performance benefits inherent to lock-free distributed communication, asynchronous systems adopting the between-graph replication pattern are more fault tolerant, as any worker can fail while the rest continue unaffected. They also exhibit a reduced memory footprint, as the replica subgraph is not replicated on the master service for each device. This is of particular importance for very large systems or models, as TensorFlow graphs serialized as protocol buffers have a maximum size of 2GB. To address these scenarios we include the \texttt{MultiWorkerAsyncReplicator}, which implements the standard between-graph parameter server architecture.

\subsection{Defining Replicas}
\label{ss:defining}
To define a replica, users define an \texttt{input\_fn} (input pipeline for the replica) and a \texttt{step\_fn} (work performed by the replica) in their TensorFlow client.

The \texttt{input\_fn} is a Python function that builds the input pipeline for a single model step, and can return either a \texttt{tf.data.Dataset} or a callable that returns an arbitrary nested structure of tensors.

\begin{python}
# TF-Replicator input function.
def input_fn(replica_id):
    def dequeue_trajectories():
        # ... code to retrieve trajectory batch.
    return dequeue_trajectories
\end{python}

TF-Replicator supports two types of input pipeline replication: \texttt{PER\_WORKER} and \texttt{PER\_REPLICA}. In the former, the datasets or callables produced by a single \texttt{input\_fn} instance are split across replicas according to an \texttt{input\_split\_fn}, which defaults to partitioning on the leading (batch) dimension. An example of where this function is useful to override is for batch-parallelism over time-major sequences to be used with recurrent networks. In the latter, each replica (e.g. GPU or TPU core) maintains its own copy of the input pipeline.

The \texttt{step\_fn} is a Python function that builds the dataflow graph for one step of one replica. It ingests the data emitted by the \texttt{input\_fn} and is executed on the targeted accelerator device (e.g. GPU or TPU). The \texttt{step\_fn} output can be any nested structure of tensors and is more general than existing solutions as it allows the execution of arbitrary TensorFlow, e.g. multiple losses and optimizers:
{\renewcommand{\arraystretch}{1.2}%
\begin{table*}[t]
\small
\centering
\begin{tabular}{l|c|c|c|c}
\multicolumn{1}{c|}{\textbf{Implementation}} & \multicolumn{1}{l|}{\textbf{Maximum workers}} & \multicolumn{1}{l|}{\textbf{Max devices per worker}} & \multicolumn{1}{l|}{\textbf{Synchronous}} & \multicolumn{1}{l}{\textbf{Parameter Servers}} \\ \hline
\texttt{NonReplicator}                                & 1                                             & 1                                                & N/A                                       & No                                             \\
\texttt{MultiGpuReplicator}                           & 1                                             & All available                                    & Yes                                       & No                                             \\
\texttt{MultiWorkerReplicator}                        & Unlimited                                     & All available                                    & Yes                                       & No                                             \\
\texttt{MultiWorkerAsyncReplicator}                   & Unlimited                                     & 1                                                & No                                        & Yes                                            \\
\texttt{TpuReplicator}                                & TPU Pod                                    & 8 TPU cores                                      & Yes                                       & No                                            
\end{tabular}
\caption{Replicator implementations for supporting data parallelism across different system architectures.}
\label{table:replicators}
\end{table*}}
\begin{python}
# Multi-optimizer TF-Replicator step function.
def step_fn(trajs):
    critic_loss = d4pg.critic_loss(trajs)
    actor_loss = d4pg.actor_loss(trajs, critic_loss)
    critic_op = critic_optimizer.minimize(critic_loss)
    actor_op = actor_optimizer.minimize(actor_loss)

    with tf.control_dependencies([critic_op, actor_op]):
        losses = tf.tuple(critic_loss, actor_loss)
    return losses
\end{python}

\subsection{Model Parallelism}
\label{ss:advanced}
One motivation of TF-Replicator is to provide general support over both data and model parallel workloads. To accomplish this, we extend the notion of a replica to support dataflow graphs that themselves span multiple devices, e.g. as necessary for models with very large memory footprints~\cite{krizhevsky2012imagenet}. This is achieved by introducing a \texttt{logical\_device} function responsible for mapping logical device IDs within the replica \texttt{step\_fn} to global device contexts. Such model parallelism can be combined with the data parallel features described above, such as this example of deploying 4 data-parallel instances of a 2-device model-parallel replica:

\begin{python}
# Combining model and data parallelism.
repl = tf_replicator.MultiGpuReplicator(
    num_gpus=8, num_gpus_per_replica=2
)
# ... code to build model and optimizer(s).
def step_fn(inputs):
    y = inputs + 1  # on GPU:0.
    with repl.logical_device(1):
        return inputs * x  # on GPU:1.
\end{python}

\subsection{Communication Primitives}
\label{ss:primitives}
Although replicated computation lends itself to most of our users' machine learning workflows, there are use cases where communication between replicas is necessary. To address these scenarios, TF-Replicator supports MPI-style primitives (similar to \texttt{torch.distributed}) that can be used to implement bespoke cross-replica reductions. Specifically, TF-Replicator implements: \texttt{all\_reduce}, \texttt{all\_sum}, \texttt{all\_gather}, \texttt{broadcast}, \texttt{map\_reduce} and \texttt{map\_gather}. The most obvious example requiring such cross-replica communication is the aggregation of gradients, as implemented in the \texttt{wrap\_optimizer} function. We achieve this in a similar manner to Horovod's \texttt{DistributedOptimizer}~\cite{sergeev2018horovod}, by leveraging TF-Replicator's \texttt{all\_sum} primitive within the optimizer's \texttt{apply\_gradients} function:

\begin{python}
# Cross-replica gradient aggregation.
def apply_gradients(self, grads, *args, **kwargs):
    num_repls = self._repl.num_replicas
    mean_grads = []
    for grad, var in grads:
        label = # ... a unique label for this gradient.
        grad = self._repl.all_sum(grad / num_repls, label) 
        mean_grads.append((grad, var))
        
    return self._optimizer.apply_gradients(
        mean_grads, *args, **kwargs
    ) 
\end{python}

All communication primitives are exposed to the user, allowing for the implementation of models that require cross-replica statistics. For example, the \texttt{all\_sum} primitive can be used to perform cross-replica batch normalization:

\begin{python}
# Cross-replica batch normalization.
def batch_norm(h):
    mean = tf.reduce_mean(h)
    mean = repl.all_sum(mean / repl.num_replicas)
    mean_sq = tf.reduce_mean(h ** 2)
    mean_sq = repl.all_sum(mean_sq / repl.num_replicas)

    return (h - mean) / tf.sqrt(mean_sq - mean)
\end{python}

Note that not all primitives are supported by the \texttt{MultiWorkerAsyncReplicator}.

\section{Implementation}
\label{sec:implementation}
TF-Replicator users describe computation in terms of the work performed by a single replica, including any communication between replicas (e.g. gradient accumulation). The implementation of such communication in TensorFlow is a non-trivial problem, as we allow the user to call these primitives anywhere within their Python \texttt{step\_fn}. If these primitives were only allowed before or after the \texttt{step\_fn} was called, i.e. while there is a global view of all replica's inputs or outputs, then cross-replica communication could be trivially stitched together. Instead, one replica's \texttt{step\_fn} can call a primitive mid graph construction. Because TensorFlow constructs dataflow graphs in a feedforward manner, this requires referring to data coming from another replica that itself is yet to be built.

The details of how computation is distributed across devices, and how communication primitives are implemented, depends on the specific Replicator instance (see Table~\ref{table:replicators}). The simplest case is \texttt{MultiWorkerAsyncReplicator}, which only supports gradient accumulation between replicas but no other communication primitives. Each worker builds its own graph according to the user-provided \texttt{step\_fn}. Asynchronous gradient accumulation is implemented by placing trainable variables on parameter servers that are shared across workers, as per the standard parameter server model~\cite{li2014scaling}. This corresponds to the between-graph replication pattern described in Section~\ref{ss:replicas}.

The \texttt{MultiGpuReplicator} and \texttt{MultiWorkerReplicator} classes implement the in-graph replication pattern. We therefore need to build a TensorFlow graph containing multiple instances of the user-provided \texttt{step\_fn} (i.e. one per replica) and handle cross-replica communication in the dataflow graph itself. One solution to the problem of feedforward graph construction in TensorFlow is to build all replicas in parallel in separate Python threads, inserting barriers where cross-replica communication is required (i.e. as implemented in TensorFlow's \texttt{DistributionStrategy}). Instead, we chose to insert placeholder ops into each graph at construction time where cross-replica communication is required. This allows each replica to be constructed independently and without the complexities of TensorFlow thread safety, as the placeholders can be rewritten once all replica subgraphs are finalized. 

The \texttt{TpuReplicator} also uses the in-graph replication across multiple TPU devices, and relies on Google's Accelerated Linear Algebra (XLA) compiler~\cite{xla} to implement cross-replica communication (i.e. using the \texttt{tf.contrib.tpu.cross\_replica\_sum} op).



\section{Case Study: Image Classification}
\label{sec:imagenet}
\begin{figure*}
\label{fig:imagenet}
\centering
\includegraphics[width=17cm]{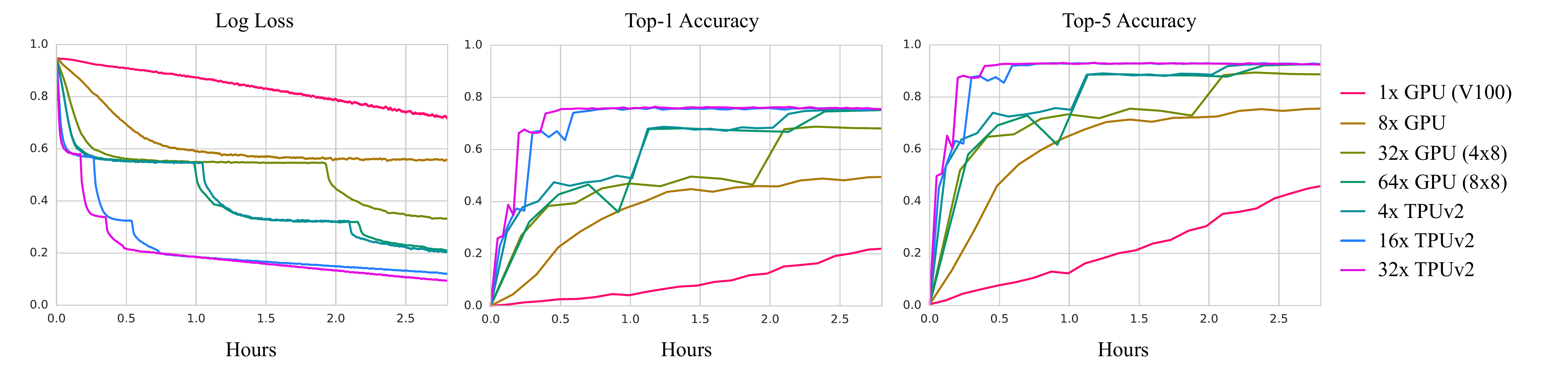}
\caption{TF-Replicator ImageNet classification performance that results from scaling a ResNetv1-50 model across different devices. Replicators used: \texttt{MultiGpuReplicator} (1, 8 GPUs), \texttt{MultiWorkerReplicator} (32, 64 GPUs), \texttt{TpuReplicator} (TPU v2).}
\end{figure*}
Since the celebrated victory of AlexNet at the 2012 Large Scale Visual Recognition Challenge (ILSVRC)~\cite{krizhevsky2012imagenet}, ImageNet classification has become a standard benchmark for evaluating deep convolutional neural networks (ConvNets). Successive ILSVRC winners such as GoogLeNet~\cite{szegedy2015going} and VGG~\cite{simonyan2014very} dramatically improved classification performance, reducing Top-5 error rate from $\sim$15\% to $\sim$5\%. The ILSVRC15-winning Residual Neural Network (ResNet~\cite{he2016deep}) introduced the concept of skip connections, allowing ConvNets many hundreds or thousands of layers deep to be trained successfully. A 152-layer ResNet exceeded human-level ImageNet performance with a Top-5 error rate of 3.57\%, and the slightly shallower ResNet-50 architecture has since emerged as the gold standard for image classification problems.

Instead of attempting to improve classification accuracy, many recent papers have focused on reducing the time taken to achieve some performance threshold (typically $\sim$75\% Top-1 accuracy). Aside from the growing body of literature on low-precision training~\cite{de2018high,micikevicius2017mixed}, these studies have typically leveraged weak scaling to train in fewer steps with very large batches.~\cite{goyal2017accurate} trained a ResNet-50 in 1 hour with batch size of 8K on 256 GPUs. Subsequent studies~\cite{codreanu2017scale,smith2017don,you2018imagenet} reduced this time to 15 minutes through a combination of larger batches and more GPUs (e.g. batch size 32K with 1024 Tesla P100s in~\cite{akiba2017extremely}), along with algorithmic tricks such as batch normalization and learning rate scheduling. \cite{baidu} introduced the idea of a ring-based all-reduce for faster cross-replica gradient updates, and similar methods have been applied successfully by~\cite{sergeev2018horovod} and~\cite{jia2018highly}, who reduced training time to 6.6 minutes using 2048 Tesla P40 GPUs.

In this Section we demonstrate how TF-Replicator can be used to obtain competitive performance on these ImageNet benchmarks, using many GPUs or TPUs in an in-graph replicated system to weak-scale a ResNet-50 ConvNet.

\subsection{Performance}
\label{ss:imagenet_performance}
We use TF-Replicator to implement a ResNetv1-50 model matching that benchmarked in the Cloud TPU documentation~\cite{cloud_tpu,goyal2017accurate}. Each TPU core or GPU receives a batch size of 64 images (32 for the largest TPU configuration, 32x TPUv2). We use a Nesterov momentum optimizer~\cite{sutskever2013importance} with a learning rate that ``warms up" linearly over 6 epochs to a maximum of $\mathrm{batch\_size} / 2048$, followed by 90\% decay at epochs 30, 60 and 80. Note that this differs from the Cloud TPU example due to our use of cross-replica gradient averaging (versus gradient summing in the TensorFlow \texttt{TPUEstimator}). Importantly, our GPU and TPU implementations use the exact same code except for the targeting of the \texttt{MultiGpuReplicator} and \texttt{TpuReplicator} respectively, each with a \texttt{PER\_WORKER}-replicated input pipeline (note that there is one TPU ``host" per device, i.e. per 4 chips / 8 cores).

Our ImageNet classification results are presented in Figure 2, in terms of the standard log loss, Top-1 and Top-5 accuracy measures. The final Top-1 accuracy and training time for 90 epochs is additionally presented in Table~\ref{table:imagenet}. Consistent with previous studies, weak-scaling from 1 to many GPU or TPU accelerators provides dramatic improvement in training time. TF-Replicator additionally allows us to deploy this model on Cloud TPU hardware, and on 32 TPUv2 devices we are able to match the published 75.3\% Top-1 accuracy in less than 30 minutes of training. Note that these results are obtained using the standard TF-Replicator implementation, without any systems optimization specific to ImageNet classification.
{\renewcommand{\arraystretch}{1.2}%
\begin{table}[t]
\small
\centering
\begin{tabular}{l|c|c}
\multicolumn{1}{c|}{\textbf{Device}} & \multicolumn{1}{l|}{\textbf{Top-1 Accuracy}} & \multicolumn{1}{l}{\textbf{Time (min)}} \\ \hline
1x GPU        & 76.19\% & 7118 \\
8x GPU        & 76.14\% & 1049 \\
32x GPU (4x8) & 76.46\% & 330  \\
64x GPU (8x8) & 76.14\% & 182  \\
4x TPUv2      & 76.34\% & 188  \\
16x TPUv2     & 75.64\% & 47   \\                                 
32x TPUv2     & 75.62\% & 28   \\
\end{tabular}
\caption{Top-1 ImageNet classification accuracy and training time after 90 epochs. Published ResNetv1-50 Top-1 accuracy is 75.3\%.}
\label{table:imagenet}
\end{table}}
\section{Case Study: Generative Models}
\label{sec:gans}
Generative models can be broadly characterized as models which learn to produce samples from a target distribution. The ability to directly model and sample from complex, high-dimensional distributions is an important task in machine learning, and permits applications ranging from speech synthesis~\cite{oord2016wavenet} to image super-resolution~\cite{ledig2017photo}. For modeling images, likelihood-based methods~\cite{dinh2016density,kingma2013auto,oord2016pixel} and implicit models~\cite{goodfellow2014generative,li2015generative,mohamed2016learning}  are the dominant approaches.

Of particular recent interest are Generative Adversarial Networks (GANs)~\cite{goodfellow2014generative} which frame the training process as a two-player minmax game wherein one network, the \textbf{generator}, attempts to produce realistic samples that fool the other network, the \textbf{discriminator}, which must learn to distinguish real and fake samples. Formally, the vanilla GAN objective, $V(G,D)$, is defined as:
\begin{equation*}
	V = \mathbb{E}_{x\sim q_{\mathrm{data}}({\mathbf{x}})} [ \log D({\mathbf{x}})] + \mathbb{E}_{{\mathbf{z}}\sim p({\mathbf{z}})} [\log(1-D(G({\mathbf{z}})))],
\end{equation*}
where typically $z \sim \mathcal{N}(0, I)$, and $G$ and $D$ are ConvNet trained through alternating gradient descent.

GANs are notoriously difficult to train, but with a recent torrent of research into stabilizing the process (some focused on modifying the objective~\cite{arjovsky2017wasserstein}, others focused on improving the network conditioning~\cite{gulrajani2017improved}) the community has settled on a reasonably robust setup, enabling training on even complex, multimodal datasets such as ImageNet. A particularly promising variant is Spectral Normalization GAN (SN-GAN)~\cite{miyato2018spectral}, which stabilizes training by normalizing each discriminator parameter by its top singular value, ensuring Lipschitz continuity and promoting convergence.

In this Section, we benchmark a class-conditional SN-GAN trained on ImageNet. We leverage TF-Replicator to train on much larger batches than can fit on a single GPU, and find that this leads to sizable gains in sample quality (as measured by the Inception Score~\cite{salimans2016improved} and Fr\'echet Inception Distance~\cite{heusel2017gans}). Without any adaptations such as changing the learning rate, simply increasing the batch size from 64 (as in~\cite{miyato2018spectral}) to 512 improves the Inception Score by a relative $\sim$50\%.
\begin{figure*}[t]
\label{fig:gan}
\centering
\includegraphics[width=17cm]{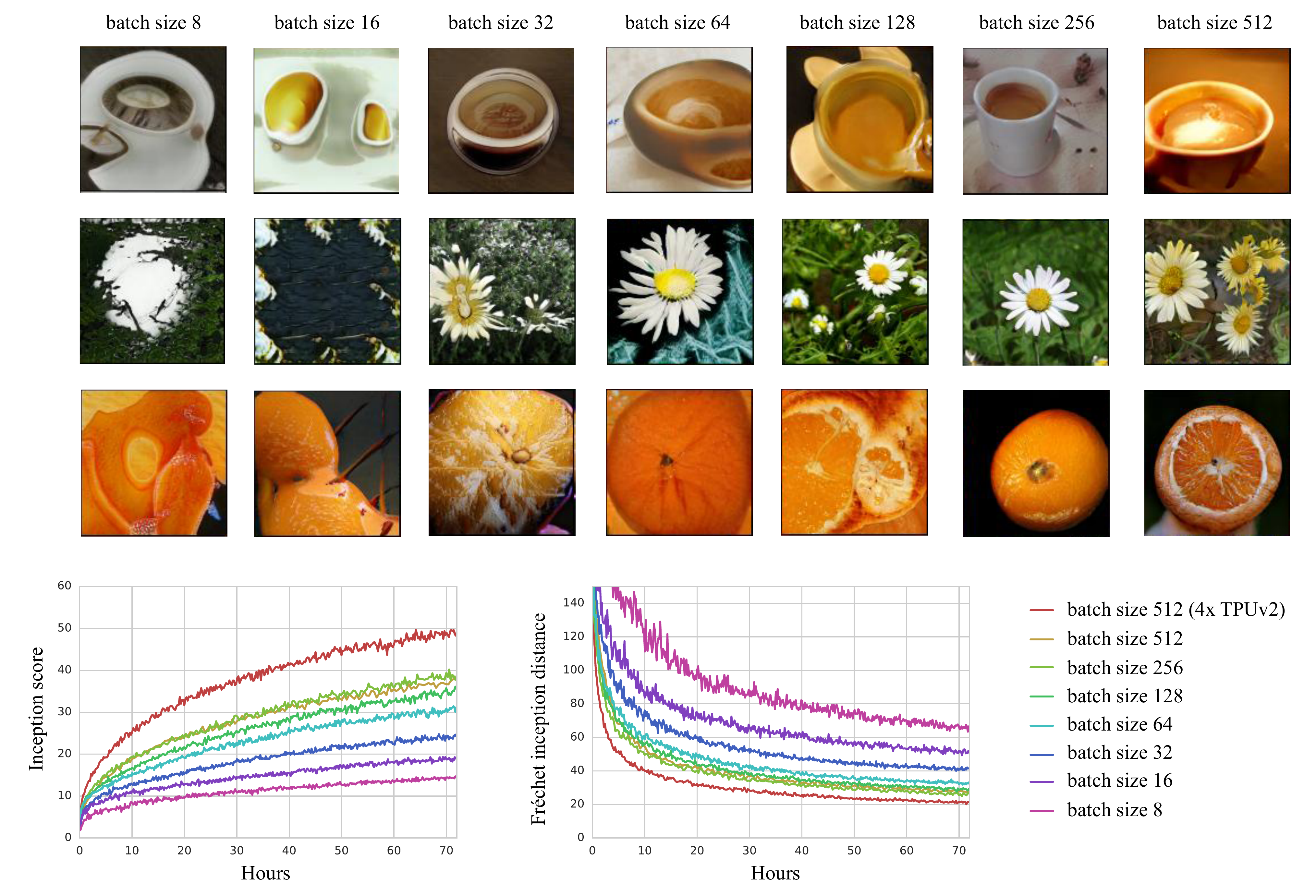}
\caption{Sample quality (Inception score and FID) and example samples of a class-conditional SN-GAN trained on 128x128 ImageNet samples. All models were trained using 8 NVIDIA Tesla V100 GPUs unless otherwise stated.}
\end{figure*}

\subsection{Performance}
\label{ss:gan_performance}
We implement SN-GAN using TF-Replicator using the same hyperparameters as~\cite{zhang2018self}. Class conditioning is provided via class-conditional BatchNorm~\cite{dumoulin2017learned} and Projection Discrimination~\cite{miyato2018cgans}. As TF-Replicator is agnostic to the number of losses defined for a model, the alternating generator and discriminator optimization steps required no additional engineering effort. We adopt the weak-scaling approach used for ImageNet classification, scaling large batches across multiple GPU and TPU devices. Our implementation further benefits from the application of cross-replica BatchNorm (implemented with TF-Replicator's \texttt{all\_sum} primitive), yielding a further performance boost of 5-10\%.

Our SN-GAN sample quality results are presented in Figure 3. By leveraging 8 GPUs (or a set of TPUv2 devices) we are able to scale to larger batches than can be fit on a single GPU, yielding substantial improvement in sample quality.

\section{Case Study: Reinforcement Learning}
\label{sec:d4pg}
\begin{figure*}[t]
\label{fig:d4pg}
\centering
\includegraphics[width=17cm]{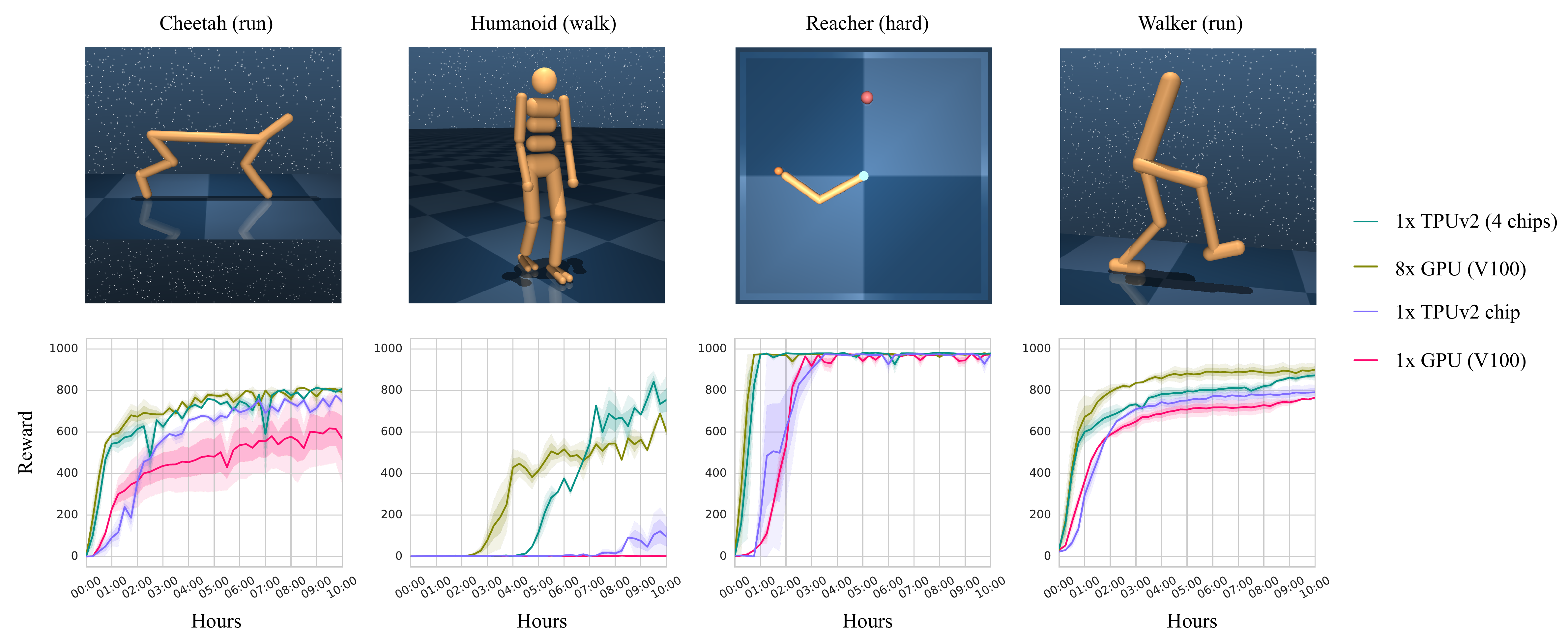}
\caption{TF-Replicator D4PG strong-scalability performance (total environment reward) on various DeepMind Control Suite tasks~\cite{tassa2018deepmind}, trained from pixel observations with a fixed total batch size of 256. Results presented for: (green) a single TPUv2 device (4 chips, 8 cores); (yellow) 8x NVIDIA Tesla V100 GPUs; (blue) 1 TPUv2 chip (2 cores); and (red) a single V100 GPU.}
\end{figure*}
In a standard reinforcement learning (RL) setup, an agent seeks to maximize the return ($\gamma$-discounted sum of future rewards) accumulated by interacting with an environment in discrete time. At each timestep $t$ the agent makes observation $\mathbf{x}_t \in \mathcal{X}$, takes actions $\mathbf{a}_t \in \mathcal{A}$ and receives rewards $r(\mathbf{x}_t, \mathbf{a}_t) \in \mathbb{R}$. The behavior of the agent is controlled by a policy $\pi : \mathcal{X} \to \mathcal{A}$, which maps each observation to an action. The state-action value function, $Q_\pi$ describes the expected return conditioned on first taking an action $\mathbf{a} \in \mathcal{A}$ from state $\mathbf{x} \in \mathcal{X}$ and following the policy $\pi$ thereafter:
\begin{equation*}
    Q_\pi(\mathbf{x}, \mathbf{a}) = \mathbb{E}\left[ \sum_{t=0}^{\infty}\gamma^t r(\mathbf{x}_t, \mathbf{a}_t)\right],
\end{equation*}
where $\mathbf{x}_0 = \mathbf{x}$, $\mathbf{a}_0 = \mathbf{a}$, $\mathbf{x}_t \sim p(. |\mathbf{x}_{t-1}, \mathbf{a}_{t-1} )$ and $\mathbf{a}_t = \pi(\mathbf{x}_t)$. In the scenario of discrete actions (e.g. Atari~\cite{mnih2015human}), the optimal policy can be derived by maximizing $Q$ with respect to $a$, where $Q$ is learned via fixed-point iteration of a Bellman operator. 

RL is complicated by continuous actions (e.g. robotics), with many successful algorithms instead learning a parameterized policy $\pi_\theta$ directly by optimizing $J(\theta) = \mathbb{E} \left[ Q_{\pi_\theta}(\mathbf{x}, \pi_\theta(\mathbf{x}))\right]$. In the case of the deterministic policy gradient (DPG) agent~\cite{lillicrap2015continuous,silver2014deterministic}, the gradient of this objective is approximated as:
\begin{equation*}
    \nabla_\theta J(\theta) \approx \mathbb{E}_\rho\left[ \nabla_\theta\pi_\theta(\mathbf{x}) \nabla_\mathbf{a}Q_{w}(\mathbf{x}, \mathbf{a}) |_{\mathbf{a}\sim\pi_\theta(\mathbf{x})} \right],
\end{equation*}
where an actor network parameterizes the policy, $\pi_\theta$, and a critic network approximates the true value of the current policy, $Q_w(\mathbf{x}, \mathbf{a}) \approx Q_{\pi_\theta}(\mathbf{x}, \mathbf{a})$. Note that similar to our GAN example, this use case requires us to support optimization over multiple losses within the \texttt{step\_fn}. Additionally, we do not require the state-visitation distribution, $\rho$, to be associated with the current policy. This allows us to decouple the process of acting (following a stochastic policy to encourage exploration) from learning the optimal deterministic policy. We accomplish this using a replay buffer of experiences, as introduced in DQN~\cite{mnih2015human}.

In this Section we describe and benchmark an implementation of the D4PG agent~\cite{barth2018distributed} on the DeepMind Control Suite~\cite{tassa2018deepmind}, a standardized set of physically-realistic control tasks of varying difficulty. D4PG is an off-policy actor-critic algorithm that extends DPG with a set of recent algorithmic advancements, e.g. a distributional Bellman operator. Importantly, unlike the benchmarks provided in the D4PG paper, the scalability of TF-Replicator allows us to quickly solve these tasks purely from pixel observations rather than relying on a low-dimensional representation of joint positions and velocities.

\subsection{Implementation}
\label{ss:d4pg_implementation}
Decoupling the processes of acting and learning is key to distributed RL agents, as both components can be independently and horizontally scaled. First, many independent \textbf{actor} processes can be deployed in parallel, each following an exploration policy to generate observation and reward statistics that can be added asynchronously to a global memory store (e.g. a replay buffer in the off-policy case). Previous RL agents built on distributed TensorFlow (e.g. IMPALA~\cite{espeholt2018impala} and ApeX~\cite{horgan2018distributed}) achieve this within a monolithic TensorFlow graph, with careful application of device scopes and variable pinning to enforce the intended behavior. Second, the \textbf{learner} process can be scaled in much the same way as a standard supervised learning workflow by performing SGD updates over larger batches of sampled actor data.

Our implementation of D4PG differs from the published version. Instead of an in-graph replay buffer, our replay is implemented out-of-graph as a memcached-style key-value store. Many independent actor processes interact with their own instance of the Control Suite environment to asynchronously generate transitions and write them to a replay buffer. The learner is implemented as a TF-Replicator replica, with additional processes responsible for sampling batches from replay and exposing a callable compatible with the replica \texttt{input\_fn}. This allows us to easily data-parallelize batches using any of the architectures or Replicators described in previous sections. The learner periodically serves out-of-graph variable update requests from RPC clients maintained by each actor, ensuring that they follow a policy centred on the most recent model parameters.

\subsection{Performance}
\label{ss:d4pg_performance}
Unlike the ResNet-50 and SN-GAN examples above, D4PG is more difficult to scale as larger batch sizes do not necessarily lead to better performance (in fact for many tasks in the Control Suite, we observe that larger batches have a negative impact). We instead use this opportunity to assess TF-Replicator's performance in scenarios requiring strong-scalability, i.e. a fixed total batch (256, the largest we could fit on a single GPU) split across multiple devices.

Consistent with the pixel benchmarking described in~\cite{tassa2018deepmind}, we consider 84x84 RGB observations representing the default camera for each task. Noting that previous studies have learnt from low-dimensional proprioceptive state that captures both the position and velocity of each actuator, we additionally frame-stack $n = 3$ frames to allow the inference of velocity information. Our agent uses the same actor and critic network architectures described in~\cite{barth2018distributed}, and the input to each is the length-64 embedding produced by a shared 9-layer ResNet over the frame-stacked pixel observations. For each experiment we use 32 actor and 8 sampler processes per learner replica to maintain a roughly equivalent acting/learning ratio (to prevent over-fitting to stale data). All other hyperparameters are as per~\cite{barth2018distributed}.

Figure 4 presents our D4PG agent performance (mean episode reward) as a function of wall-clock training time. It is evident that in this strong-scalability regime and with a modest batch size of 256, TF-Replicator allows for substantial acceleration of agent training time. Consistent with our ResNet-50 and SN-GAN weak-scalability results, we observe that a single TPUv2 device (8 cores across 4 chips) provides competitive performance compared to 8 NVLink-connected Tesla V100 GPUs.

\section{Related Work}
\label{sec:related}
Early systems for distributed machine learning built upon the popular MapReduce batch dataflow architecture~\cite{dean2008mapreduce}. MapReduce adopts the functional programming $\texttt{map(k1, v1)} \to \texttt{list[k2, v2]}$ and $\texttt{reduce(k2, list[v2])} \to \texttt{list[v2]}$ primitives, and provides a fault tolerant architecture that could scale to many thousand machines. Examples of MapReduce-based systems for machine learning include Mahout~\cite{mahout}, based on Hadoop~\cite{shvachko2010hadoop}, and MLI~\cite{sparks2013mli}, based on Spark~\cite{zaharia2012fast}. Although these systems provided early success when scaled to tens of machines, the MapReduce paradigm is relatively ill-suited to scaling the iterative computation of deep neural networks~\cite{abadi2016tensorflow}. For example, the SparkNet system (based on Spark) requires 20 seconds to broadcast weights and collect updates from just 5 workers, despite extending earlier frameworks to store intermediate results in memory~\cite{moritz2015sparknet}.

Subsequent systems have extended the parameter server model, introduced by~\cite{smola2010architecture} and later popularized by~\cite{li2013parameter,li2014scaling} for machine learning. This architecture separates tasks into two jobs: stateless workers, which perform model computations; and stateful parameter servers, which store sharded network weights. The parameter server itself was inspired largely by memcached~\cite{fitzpatrick2004distributed}, exposing \texttt{push(k,v)} and \texttt{pull(k)} functions that can be used for applying gradient-based updates to shared neural network weights. Examples of parameter server-based systems for machine learning include DistBelief~\cite{dean2012large}, GeePS~\cite{cui2016geeps} and Project Adam~\cite{chilimbi2014project}. A particularly noteworthy example is MXNet~\cite{chen2015mxnet}, which uses dataflow graphs to represent worker computation in a similar manner to TensorFlow~\cite{abadi2016tensorflow}.

More recent frameworks include Mesh-TensorFlow~\cite{shazeer2018mesh}, which defines a new language that simplifies data and model-parallelism by allowing the user to explicitly map sub-computations to mesh-connected hardware; and Ray~\cite{moritz2018ray}, which unifies simulation, training and serving for distributed reinforcement learning models. Both frameworks have demonstrated impressive scalability performance, e.g. Mesh-TensorFlow achieves state-of-the-art English-to-French translation by training a 5-billion parameter Transformer model on 512 TPU cores. TF-Replicator instead aims to promote rapid iteration of research ideas, allowing researchers to scale their existing, single-machine models to a distributed system without the need to learn new tools or reason explicitly about network topology.

\section{Conclusion}
\label{sec:discussion}
We have presented TF-Replicator, and demonstrated that it is a useful abstraction that simplifies the process of scaling general machine learning models. Without any distributed systems expertise, the user is able to leverage the power of clusters of GPUs or TPUs to achieve results competitive with complex hand-crafted systems. We believe that massive scalability will continue to be an important ingredient of machine learning research, and that TF-Replicator will be a useful tool for enabling impactful outcomes in this domain.

\section*{Contributions}
TF-Replicator design: F.B., P.B., D.G., C.J., F.V.; TF-Replicator implementation: P.B., C.J.; Experimental design and paper writing: D.B.; ImageNet experiments: S.G.C.; D4PG experiments: J.A., D.B., A.C.; GAN experiments: A.B.; General support: D.B., D.G., A.P.

\section*{Acknowledgements}
Thanks to Tim Harley, Tom Hennigan, Karen Simonyan, Koray Kavukcuoglu and the entire DeepMind research community.

\bibliography{replicator}
\bibliographystyle{sysml2019}

\end{document}